\documentclass[cameraready]{Interspeech}
\usepackage{tipa}
\usepackage[utf8]{inputenc}
\usepackage{enumitem}
\usepackage{hyperref}

\usepackage{booktabs}

\usepackage{ragged2e} 
\usepackage{tabularx}

\title{Speaker or Language? Explaining Variance in Charismatic Prosody Across Luxembourgish and French}

\author[
    affiliation={1,2},
    orcid=0000-0002-0821-9125,
    correspondingauthor
]{Nina}{Hosseini-Kivanani}

\author[
    affiliation={3},
    orcid=0000-0001-9715-4505,
]{Nafiseh}{Taghva}

\author[
    affiliation={2},
    orcid=0000-0001-9216-3082
]{Peter}{Gilles}

\author[
    affiliation={4},
    orcid=0000-0002-8623-1680
]{Oliver}{Niebuhr}

\address{
    $^1$Radio Télévision Luxembourg (RTL), Luxembourg\\
    $^2$ Faculty of Humanities, Education and Social Sciences, University of Luxembourg, Luxembourg \\
    $^3$ Department of Foreign Languages and Linguistics,
    Shiraz University, Iran  \\
    $^4$ Centre for Industrial Electronics, University of Southern Denmark, Denmark 
}

\email{nina.hosseinikivanani@ext.uni.lu,taghvanafiseh@gmail.com,peter.gilles@uni.lu,olni@sdu.dk
}


\begin{document}

\maketitle

\begin{abstract}

Charismatic speech is shaped by language and speaking style, yet their relative contribution in bilingual public speaking remains unclear. We analyzed spontaneous speeches of 10 politicians who address audiences in Luxembourgish and French, in highly comparable communicative contexts across languages. From 400 utterances, we extracted 41 acoustic–prosodic features linked to vocal charisma and fitted mixed-effects models to separate speaker- and language-related variance. Speaker identity accounted for most variance, whereas language explained less, but still showed systematic differences: French productions showed higher shimmer and phrase-final F0, indicative of a polite, respectful voice, while Luxembourgish productions exhibited stronger mid-frequency spectral energy, suggesting a more vocally present profile. These patterns align with the sociolinguistic roles of Luxembourgish as an informal identity language and French as a high-prestige institutional variety.

\end{abstract}

\section{Introduction}
The systematic study of persuasive speaking originated not in the political but in the legal practices of ancient Greece, where structured techniques were developed to convince juries in public court proceedings. It happened later that this knowhow was extended to political speeches, most notably by Aristotle, whose concept of \textit{ethos} established that persuasive impact arises not only from factual argumentation (\textit{logos}) but from the credibility and competence attributed to the speaker during the act of speaking~\cite{antonakis2016charisma}. This thousands-of-years-old line of research continues today under Max Weber's term charisma~\cite{adair2020max}. Political figures have ever since remained a main focus of charisma research. The present study focuses on political speakers, too~\cite{brecher2016political}. 

A substantial body of research in phonetics and speech communication demonstrates that charismatic impressions are systematically related to multidimensional acoustic–prosodic patterns rather than single cues~\cite{signorello2012charisma}. Foundational studies on political speech showed that charismatic speakers exhibit characteristic differences in pitch (i.e., fundamental-frequency or F0) modulation, intensity dynamics, and lexical choice~\cite{RosenbergHirschberg2005,rosenberg2009charisma}. Subsequent work has provided converging statistical evidence that higher charisma ratings are generally associated with increased F0 level and variability, expanded F0 range, enhanced F0 variability, greater intensity level and variability, and more dynamic temporal–melodic shaping of utterances~\cite{strangert2008makes,biadsy2008cross,niebuhr2018shapes,niebuhr2019measuring,niebuhr2021those}.

At the same time, several parameters show systematic negative relationships with perceived charisma. Lower phrase-final F0 values and shorter phrase durations have been shown to correlate with higher charisma ratings~\cite{niebuhr2018shapes}. In addition, a broad class of voice-quality measures related to spectral tilt, including long-term spectral slope estimates and amplitude differences such as H1–H2 or H1–A3 has been found to correlate negatively with charisma, with shallower spectral slopes and smaller amplitude differences consistently associated with higher perceived charisma~\cite{niebuhr2018acoustic}. Finally, recent work shows that filled pauses are also negatively related to perceived speaker performance and charisma, with fewer and shorter fillers contributing to more positive listener impressions~\cite{niebuhr2019not}.

On the whole, this literature converges on the view that charismatic speech is characterized by a systematic enhancement of prosodic effort and dynamics, combined with reduced hesitation and clear temporal organization, resulting in a distinct acoustic profile that is robustly perceived across contexts.

Despite this progress, most existing research has examined monolingual speakers and relied on prepared or semi-controlled speech. Much less is known about how charisma-related prosody is realized in spontaneous bilingual speech, particularly when speakers alternate between languages that differ in sociolinguistic function and prestige. Independent of charisma research, studies on bilingual prosody consistently report systematic language-dependent differences in pitch range, timing, and voice quality for the same speaker~\cite{mennen2004bi,de2008your,passoni2022bilinguals}. Such differences can reflect both structural linguistic constraints and social-indexical meanings such as formality or politeness. Voice-source characteristics and spectral distributions have likewise been shown to shift across languages in perceptually salient ways~\cite{esposito2020cross,sicoli2010shifting}. These findings strongly suggest that language choice can shape core prosodic patterns, yet its implications for charismatic speech remain largely unexplored.

Luxembourg provides a particularly informative sociolinguistic context for addressing this gap. Although officially trilingual, the functional distribution and ideological status of its languages are highly asymmetrical. Luxembourgish is strongly associated with national identity and everyday spoken interaction, whereas French dominates in administration, law, and high-prestige institutional domains~\cite{gilles2006luxembourgish,horner2016language,horner2017introducing}. Sociolinguistic research consistently characterizes French as a prestige language and Luxembourgish as a key marker of local belonging and authenticity. The third language, German, by contrast, plays an important role primarily in writing, education, and the media, but is less central to spontaneous spoken interaction. Accordingly, our study focuses on Luxembourgish and French.

This sociolinguistic configuration gives rise to two competing expectations: speakers may enhance charisma-related prosodic cues when speaking French in order to align with high-prestige norms, or they may instead display stronger, i.e. more pronounced charisma cues when speaking Luxembourgish as the language of in-group affiliation and authenticity. Thus, our first research question is as follows.

\textbf{RQ1.} Do speakers systematically realize charisma-related prosodic cues differently when speaking French compared to Luxembourgish, and if so, in which direction? Specifically, are such cues overall stronger in the prestige language (French) or in the identity language (Luxembourgish)?

Additionally, if such language-dependent differences exist, a further question concerns their magnitude relative to individual speaker differences. While previous work has demonstrated robust speaker-specific prosodic “signatures”~\cite{nolan1999speaker} across contexts, styles and even (non-)native languages, it remains unclear whether such idiosyncratic patterns outweigh potential language effects in prestige-asymmetric bilingual contexts.

\textbf{RQ2.} Under otherwise comparable communicative conditions, does language explain more variance in charisma-related prosodic features than individual speaker differences within gender groups? In other words, are the cross-language differences produced by the same speakers larger than the differences observed between speakers (of the same gender) within a given language?

We analyze spontaneous political speech produced by ten Luxembourg politicians (five female, five male) who regularly speak in both languages in public contexts. From each speaker, 20 utterances in Luxembourgish and 20 in French were extracted, yielding a parallel within-speaker dataset of 400 utterances. This design allows us to directly compare how the same speakers realize charisma-related prosody when speaking the two languages.

To ensure valid cross-language comparisons, the Luxembourgish and French samples were selected to be maximally comparable with respect to situational, interactional, and communicative context. All recordings stem from similar public, political, or institutional settings, involve comparable audiences, and reflect closely matched discourse functions. All material is spontaneous; no read or scripted speech was included. Although naturalistic data inevitably involve some contextual variability, the dataset was constructed to minimize such differences as far as possible, enabling principled within-speaker comparisons under ecologically valid conditions.


Using a set of established acoustic correlates of vocal charisma derived from prior work on political and business speech, including the recent overview by~\cite{berger2024like}, we quantify the relative contributions of speaker and language to prosodic variability in spontaneous bilingual speech and test for systematic prestige-related modulation.

\section{Methods}
\subsection{Speakers and material}

We analyzed speech from ten high-profile public figures in Luxembourg: five male and five female speakers~\footnote{The data was collected from RTL Archive with permission.}. 
All speakers are functionally bilingual in Luxembourgish and French and regularly use both languages in public communication, for example, the prime minister, ministers, and members of government or public institutions, see Table~\ref{tab:lux_politicians}.

\begin{table}[t]
  \centering
  \scriptsize              
  \setlength{\tabcolsep}{1.75pt} 
  \caption{Current roles of selected Luxembourgish politicians.}
  \label{tab:lux_politicians}
  \begin{tabularx}{\columnwidth}{l c c X}
    \toprule
    \textbf{Name} & \textbf{G} & \textbf{Age} & \textbf{Current role (Luxembourg)} \\
    \midrule
    Elisabeth Margue & F & 35 &
    Minister of Justice; Del. PM for Media, Connectivity, Parliament \\
    Lydie Polfer & F & 73 &
    Mayor of Luxembourg City \\
    Paulette Lenert & F & 57 &
    MP (LSAP); former Dep. PM and Minister of Health \\
    Taina Bofferding & F & 43 &
    MP (LSAP); President of Socialist parliamentary group \\
    Tilly Metz & F & 58 &
    MEP (Greens/EFA, déi Gréng) \\
    Claude Meisch & M & 54 &
    Minister of Education, Children and Youth; Housing and Spatial Planning \\
    Gilles Roth & M & 58 &
    Minister of Finance \\
    Jean Asselborn & M & 76 &
    Former Minister for Foreign and European Affairs \\
    Luc Frieden & M & 62 &
    Prime Minister; President of CSV \\
    Xavier Bettel & M & 52 &
    Dep. PM; Foreign Affairs and Trade; Development Cooperation \\
    \bottomrule
  \end{tabularx}
\end{table}

For each speaker, we selected spontaneous segments from public speeches, press briefings, interviews, or parliamentary debates in Luxembourgish and in French. From these materials, we then extracted
\begin{itemize}
    \item 20 spontaneous sentences in Luxembourgish per speaker
    \item 20 spontaneous sentences in French per speaker.
\end{itemize}

Sentences were defined as intonationally and syntactically coherent units. They were segmented manually based on audible prosodic boundaries, supported by punctuation in transcripts where available. The sentence contents are not parallel across languages, because the speeches were delivered in different contexts, but they are comparable in style and domain. All speech is spontaneous, not read. Segments containing overlapping talk, background noise, or prominent non-speech vocalizations (for example, laughter or coughing) were excluded from the dataset.

\subsection{Transcription, annotation, and prosodic measures}

All original speech material was first extracted from the source video files as mono audio at a sampling rate of 16 kHz. Orthographic transcriptions for both Luxembourgish and French segments were then generated automatically using the LuxASR system~\cite{gilles2023asrlux,gilles2023lux} and subsequently checked and corrected for obvious recognition errors where necessary. These transcriptions provided the basis for subsequent analyses.



Next, the temporal boundaries of the 20+20 selected sentences were refined in Praat. For each speaker–language pair, we annotated sentence boundaries. The recordings were automatically segmented and annotated at the sentence level using WebMAUS~\cite{kisler2017multilingual}, which employs language-specific phonological models for forced alignment. Then, all automatically generated boundaries were reviewed and, where necessary, manually adjusted with reference to both the waveform and a broadband spectrogram, following established phonetic segmentation criteria~\cite{pollak2008phone}. The finalized annotations for each recording were exported as Praat TextGrid files.

Acoustic features were extracted using the ProsodyPro script for Praat, which provides a systematic pipeline for large-scale prosodic analysis~\cite{Xu2013ProsodyPro}. ProsodyPro generates a range of discrete measurements suitable for statistical analysis, including F0 measures (e.g., maximum, minimum, mean F0, F0 range), intensity measures (e.g., mean intensity), duration and timing measures (e.g., interval duration), and additional interval-based metrics derived from labeled intervals.

In addition, ProsodyPro computes a suite of bio-informational dimension (BID) measures related to voice quality and spectral characteristics, such as amplitude differences between harmonics (e.g., H1–H2, H1–A3), spectral slope indicators (e.g., Hammarberg index, spectral center of gravity), cepstral peak prominence (CPP), jitter and shimmer, harmonicity (HNR), and energy distributions across frequency bands~\cite{Xu2013ProsodyPro}. These measures collectively capture multiple aspects of prosody and phonation relevant to our analysis.


\subsection{Statistical analysis}

All analyses were carried out on the 41 prosodic and acoustic features~\footnote{Link to the 41 features: \href{https://anonymous.4open.science/r/Charismatic-Prosody-58F6/README.md}{https://anonymous.4open.science}} described above, treating each of the 400 sentence tokens as a separate observation. Before modeling, each feature was inspected for outliers and obvious artifacts. Strongly skewed measures were log-transformed where necessary, and all features were then standardized to zero mean and unit variance across the full dataset. This scaling yields comparable coefficients and allows us to interpret fixed effect estimates as standard deviation differences between conditions.

To obtain a global view of the prosodic space, we first ran a principal component analysis (PCA) on the 41 z-scored features. The leading components and their loadings were descriptively used in the Results section to visualize how sentences cluster by speaker and by language, and to identify feature bundles that drive the main prosodic contrasts. The core inferential analyses addressed the two research questions using linear mixed effects models fitted separately for each feature. The baseline model was
\[
\text{feature} \sim \text{Language} + \text{Gender} + \text{Duration} + (1 \mid \text{Speaker})\,,
\]
where Language (Luxembourgish vs French), Gender (female vs male), and sentence Duration are fixed effects, and Speaker is a random intercept. From this model, we extracted the variance of the random Speaker term and the residual variance to compute an intraclass correlation coefficient for Speaker, \(\mathrm{ICC}_{\text{Speaker}}\), which reflects the proportion of total variance attributable to stable between speaker differences. The contribution of Language was estimated from the same models using marginal \(R^2\) and type III sums of squares, allowing us to compare the variance shares associated with speaker identity and language choice across the 41 features.

To address RQ1, we extended the baseline model by adding an interaction term \(\text{Language} \times \text{Gender}\). The fixed effect of Language, expressed on the z-score scale, captures the average difference between French and Luxembourgish for each feature, while the interaction tests whether this difference depends on gender. For each feature, we report the estimated Language effect, its standard error, and a \(p\) value based on Satterthwaite degrees of freedom. To account for multiple testing across the 41 features, \(p\) values for the Language main effects were adjusted using the Benjamini–Hochberg false discovery rate procedure.

\section{Results}

\subsection{RQ1. Language differences in charisma-related cues}

\begin{figure}[hpt!]
    \centering
    \includegraphics[width=1.04\linewidth, height=0.55\linewidth]{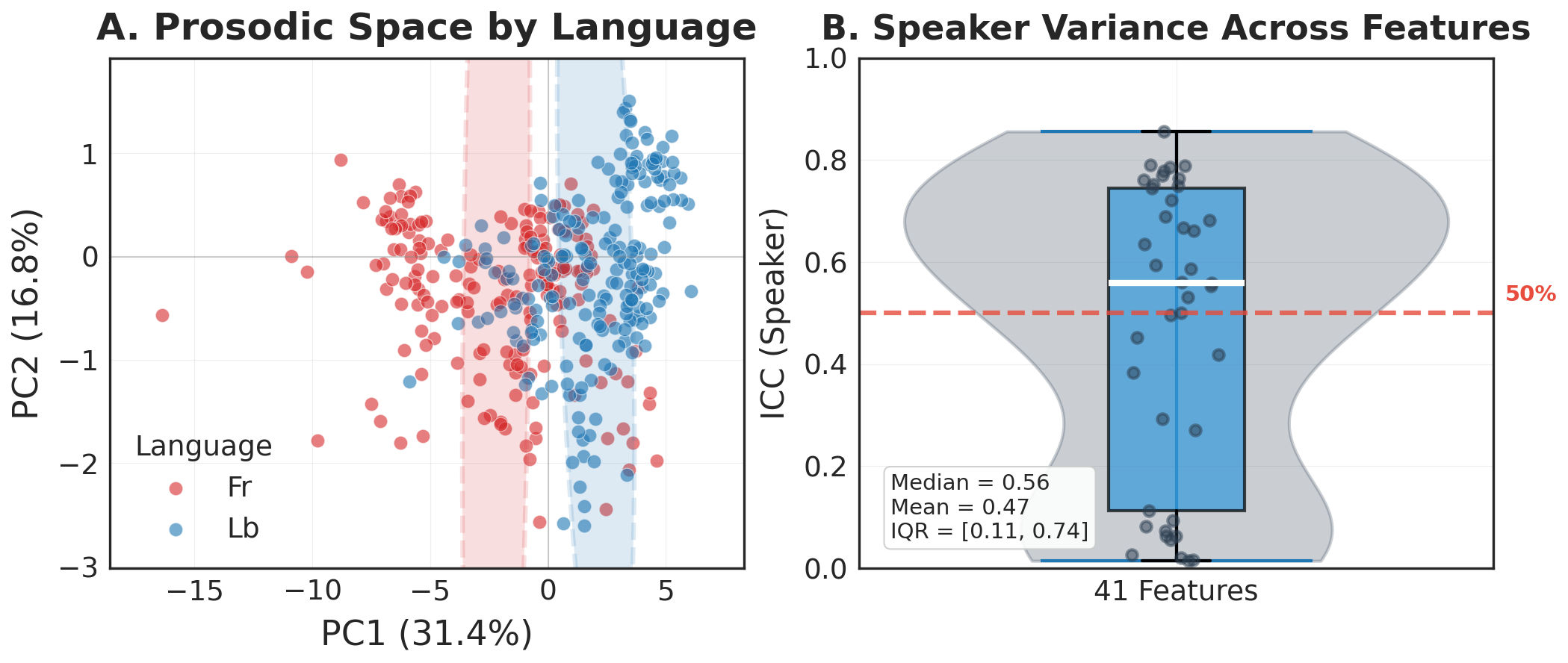}%
    \caption{
Prosodic structure and variance partitioning.
Panel~A. PCA map of sentences in prosodic space, coloured by language (Fr vs Lb).
Panel~B. Speaker intraclass correlations (ICC) across 41 features.
Together, these plots show that sentences cluster mainly by speaker, and that speaker identity explains more than half of the variance for most features.
}
    \label{fig:pca_var}
\end{figure}

The principal component analysis provides an overview of how sentences are distributed in the prosodic feature space. The first two components account for 31.4\% and 16.8\% of the total variance, respectively. In the PCA map, French and Luxembourgish sentences form partially overlapping bands along PC1, with Luxembourgish tokens shifted towards higher PC1 values but substantial overlap between languages (Figure~\ref{fig:pca_var}A). Thus, language choice induces a systematic but comparatively modest displacement in the 41 dimensional space.

The mixed effects variance partitioning confirms that speaker identity is the dominant source of variability. Across the 41 prosodic and acoustic features, the intraclass correlation for Speaker has a median of 0.56 (IQR 0.11–0.74, range 0.01–0.86), so that for a typical feature, more than half of the variance is attributable to stable differences between speakers (Figure~\ref{fig:pca_var}B). In contrast, Language accounts for a median of only 0.5\% of the variance across features. The most speaker-stable cues, including median and mean F0, the 250~Hz energy level, and several mid-frequency levels, show speaker-related variance shares around 70–80\%. 

\subsection{RQ2. Speaker versus language in prosodic variation}

\begin{figure}[t!]
    \centering
    \includegraphics[width=1.02\linewidth, height=0.52\textwidth]{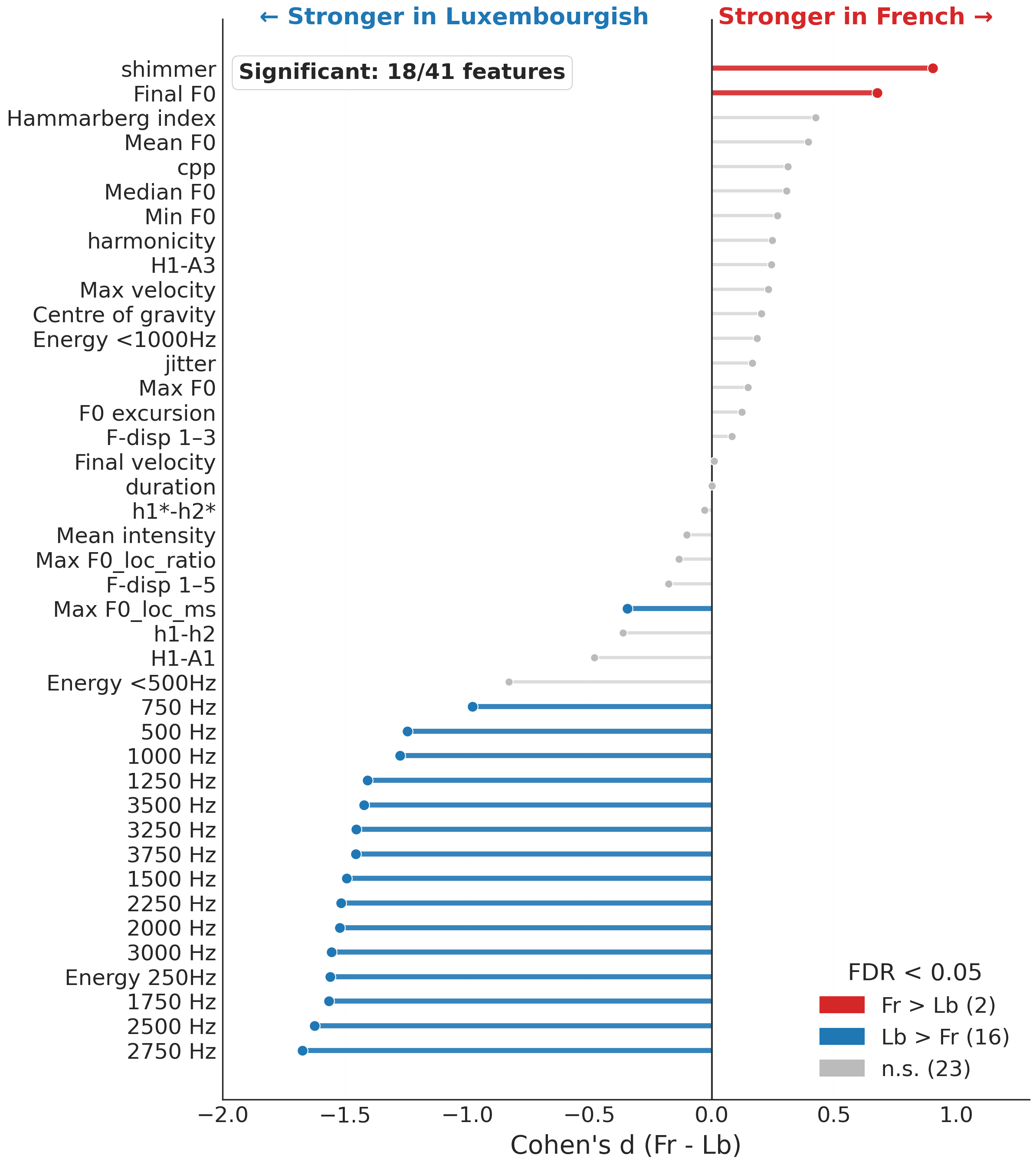}%
    \caption{Language effects across prosodic features.
    Cohen's $d$ is computed for French minus Luxembourgish.
    Blue lines mark features that are significantly stronger in Luxembourgish, red lines those significantly stronger in French, and grey points non-significant effects after FDR correction.
    See Table~\ref{tab:language-effects} for values.}
    \label{fig:lang_effects}
\end{figure}

Although language explains only a small share of total variance, several prosodic dimensions show systematic shifts between French and Luxembourgish. Mixed effects models with Language, Gender, and their interaction as fixed effects and random intercepts by Speaker reveal significant Language effects for 18 of the 41 features after Benjamini–Hochberg correction. The effect sizes are summarized in Figure~\ref{fig:lang_effects} and detailed in Table~\ref{tab:language-effects}. Only two measures are reliably higher in French. shimmer and final F0, with Cohen’s \(d = 0.90\) and \(d = 0.68\), respectively. This pattern points to somewhat higher phrase final pitch and greater amplitude perturbation in French, together with a small tendency towards clearer voice quality as reflected in positive but non-significant shifts in CPP and the Hammarberg index.

In contrast, 16 features show higher values in Luxembourgish. These include the low-frequency energy band around 250~Hz and almost all long-term spectral bands between 500 and 3750~Hz, with large negative effect sizes such as \(d = -1.67\) at 2750~Hz and \(d = -1.62\) at 2500~Hz, as well as a moderate shift in the timing of the F0 maximum. The direction and size of these effects suggest a brighter, higher-energy spectral balance in Luxembourgish, whereas French productions are comparatively spectrally damped. No Language by Gender interactions survive FDR correction, therefore, the direction of language differences is similar for male and female speakers.

To summarize these multidimensional patterns, we constructed a composite charisma index from six established cues. median pitch and excursion size, and Hammarberg index, low frequency energy, jitter, and shimmer, with feature signs aligned so that higher values correspond to more charismatic settings. A mixed effects model on this index showed no reliable main effect of Language (Cohen’s \(d = -0.03\), \(p = 0.89\), \(R^2_{\text{Language}} \approx 0\)) and no Language by Gender interaction. While individual speakers differ markedly in their overall charisma scores, their relative ordering across languages is inconsistent. Taken together with the feature-wise analysis, this indicates that French and Luxembourgish differ in specific spectral and contour properties, but these shifts do not yield a clear global charisma advantage for either language.

\begin{table}[hpt!]
\centering
\scriptsize
\caption{Prosodic features with significant language effects after FDR correction. $d$ is for French minus Luxembourgish (positive = FR $>$ LB). Band labels give the center frequency in Hz.}
\label{tab:language-effects}

{\setlength{\tabcolsep}{14pt}
 \renewcommand{\arraystretch}{0.65}

\begin{tabular}{lccc}
\toprule
Feature & Domain & Dir. & $d$ \\
\midrule
\multicolumn{4}{c}{\textit{French $>$ Luxembourgish}} \\
\midrule
Shimmer      & Voice        & FR $>$ LB & 0.90 \\
Final F0     & F0 contour   & FR $>$ LB & 0.68 \\
\midrule
\multicolumn{4}{c}{\textit{Luxembourgish $>$ French}} \\
\midrule
EProf 250    & Spectrum     & LB $>$ FR & $-1.56$ \\
Band 500     & Spectrum     & LB $>$ FR & $-1.24$ \\
Band 750     & Spectrum     & LB $>$ FR & $-0.98$ \\
Band 1000    & Spectrum     & LB $>$ FR & $-1.28$ \\
Band 1250    & Spectrum     & LB $>$ FR & $-1.41$ \\
Band 1500    & Spectrum     & LB $>$ FR & $-1.49$ \\
Band 1750    & Spectrum     & LB $>$ FR & $-1.57$ \\
Band 2000    & Spectrum     & LB $>$ FR & $-1.52$ \\
Band 2250    & Spectrum     & LB $>$ FR & $-1.52$ \\
Band 2500    & Spectrum     & LB $>$ FR & $-1.62$ \\
Band 2750    & Spectrum     & LB $>$ FR & $-1.67$ \\
Band 3000    & Spectrum     & LB $>$ FR & $-1.56$ \\
Band 3250    & Spectrum     & LB $>$ FR & $-1.45$ \\
Band 3500    & Spectrum     & LB $>$ FR & $-1.42$ \\
Band 3750    & Spectrum     & LB $>$ FR & $-1.46$ \\
F0 max time  & F0 contour   & LB $>$ FR & $-0.35$ \\
\bottomrule
\end{tabular}
}

\end{table}

\section{Discussion and Outlook}

The present study examined how language choice and speaker identity shape charisma-related prosody in spontaneous bilingual political speech. Two research questions were pursued.

\textbf{RQ1 asked whether charisma-related prosodic cues are systematically realized differently in French and Luxembourgish.} The results show language-specific differences, but primarily in F0 and voice quality characteristics rather than in global prosodic scaling. French productions exhibit higher phrase-final F0 and higher shimmer, whereas Luxembourgish productions show consistently stronger mid-frequency spectral energy and slightly earlier F0 peak timing. Importantly, the elevated phrase-final F0 in French is likely influenced by language-specific intonational phonology (e.g., high phrase-final boundary tones), and should therefore not be interpreted straightforwardly as a stylistic choice~\cite{delais2015intonational}.

Beyond phonological contributions, the systematic spectral differences point to language-dependent phonatory settings in political speech that align well with the sociolinguistic functions of the two languages. Luxembourgish, which functions as a language of informality, authenticity, and in-group affiliation, is associated with a more vocally \emph{present} and projected profile (higher mid-frequency energy; shallower spectral tilt). French, by contrast, is linked to institutional authority, formality, and prestige and tends toward a comparatively more \emph{considerate} and controlled phonatory setting, reflected in higher shimmer and tendencies toward higher F0 and steeper spectral tilt. These subtle but consistent differences suggest that speakers adjust voice-quality-related cues when switching languages in ways that index social meanings such as formality and politeness across codes~\cite{brown2017politeness}. In line with recent evidence that politeness-related stances are not necessarily cued by pitch alone but often rely on loudness and voice-quality dimensions~\cite{idemaru2020loudness}, the present pattern is compatible with a pragmatic interpretation in which French is produced with a more institutionally oriented, socially considerate voice setting. 

\textbf{RQ2 asked whether language differences outweigh speaker-specific variation.} Here, the answer is clearly negative. Variance partitioning shows that speaker identity remains by far the dominant source of prosodic variability, accounting for more than half of the variance across most features, whereas language explains less. Thus, charisma-related prosody appears primarily speaker-specific rather than language-driven. Notably, this pattern closely aligns with observations from public speaking practice: Language switching alone does not turn weak speakers into strong ones, nor vice versa~\cite{gaffney2021effects}.

Several limitations qualify these conclusions. The dataset comprises only ten high-profile speakers from a single speech community and is restricted to political speech. Moreover, the present findings are based on acoustic analyses rather than direct listener judgments. On the basis of established correlations between acoustic parameters (including voice-quality measures) and perceived speaker charisma, one would predict that the Luxembourgish productions in this dataset should, on average, be judged as slightly more charismatic than the French ones. Testing this hypothesis, however, requires dedicated perception experiments. Future work should therefore combine production and perception approaches, extend the corpus to additional speaker groups and communicative settings, and model discourse structure more explicitly to capture how language choice, style, and charismatic strategies unfold over time.

\section{Acknowledgments}
This research was conducted in the context of the LuxVoice
project, funded by the Luxembourg National Research Fund
(FNR) under grant agreement (project reference 19205922).
LuxVoice aims to advance Luxembourgish language tech-
nologies and support the development of robust multilingual
AI resources. The authors gratefully acknowledge the support of RTL Lëtzebuerg and the University of Luxembourg.
\section{Generative AI Use Disclosure}
The use of generative AI (i.e., Grammarly) in this paper was limited strictly to text refinement and clarity improvements. The authors are solely responsible for all scientific content, including the methodology, experiments, analysis, and conclusions.

\bibliographystyle{IEEEtran}
\bibliography{mybib}

\end{document}